\documentclass[runningheads]{llncs}
\usepackage{amsmath}
\usepackage{booktabs} 
\usepackage{caption} 
\usepackage{graphicx}
\usepackage{pgfplots}
\usepackage[all]{nowidow}
\usepackage[utf8]{inputenc}
\usepackage{tikz}
\usetikzlibrary{er,positioning,bayesnet}
\usepackage{algpseudocode,algorithm,algorithmicx}
\usepackage[inline]{enumitem} 

\definecolor{blue}{HTML}{1F77B4}
\definecolor{orange}{HTML}{FF7F0E}
\definecolor{green}{HTML}{2CA02C}

\pgfplotsset{compat=1.14}

\begin{document}
	\title{Towards Early Diagnosis of Epilepsy from EEG Data}
	%
	%
	\author{Diyuan Lu\inst{1, 2, 3} \and
		Sebastian Bauer\inst{2,3} \and
		Valentin Neubert\inst{4} \and
		Lara Sophie Costard\inst{5} \and
		Felix Rosenow\inst{2,3} \and
		Jochen Triesch\inst{1, 2, 3}}
	%
	\institute{Frankfurt Institute for Advanced Studies (FIAS), Frankfurt am Main, 60438, Germany
		\email{\{elu, triesch\}@fias.uni-frankfurt.de}\\ \and
		Goethe University Frankfurt, 60438 Frankfurt am Main, Germany\\ \and
		Center for Personalized Translational Epilepsy Research (CePTER), Frankfurt am Main, Germany\\
		\email{Sebastian.Bauer@kgu.de} \\ 
		\email{rosenow@med.uni-frankfurt.de}\\ \and
		Oscar Langendorff Institute of Physiology, Rostock University Medical Center, Rostock, Germany \\
		\email{valentin.neubert@uni-rostock.de}\\ \and
		Tissue Engineering Research Group, Royal College of Surgeons Ireland, St Stephans Green 123, Dublin, D02, Ireland\\
		\email{laracostard@rcsi.com}}
	\maketitle              
	\begin{abstract}
		Epilepsy is one of the most common neurological disorders, affecting about 1\% of the population at all ages. Detecting the development of epilepsy, i.e., epileptogenesis (EPG), before any seizures occur could allow for early interventions and potentially more effective treatments. Here, we investigate if modern machine learning (ML) techniques can detect EPG from intra-cranial electroencephalography (EEG) recordings prior to the occurrence of any seizures. For this we use a rodent model of epilepsy where EPG is triggered by electrical stimulation of the brain. We propose a ML framework for EPG identification, which combines a deep convolutional neural network (CNN) with a prediction aggregation method to obtain the final classification decision. Specifically, the neural network is trained to distinguish five second segments of EEG recordings taken from either the pre-stimulation period or the post-stimulation period. Due to the gradual development of epilepsy, there is enormous overlap of the EEG patterns before and after the stimulation. Hence, a prediction aggregation process is introduced, which pools predictions over a longer period. By aggregating predictions over one hour, our approach achieves an area under the curve (AUC) of 0.99 on the EPG detection task. This demonstrates the  feasibility of EPG prediction from EEG recordings.
		
	\end{abstract}
	\section{Introduction}
	
	Identifying patients at high risk of developing epilepsy is of great importance to allow early medical intervention and improve the effectiveness of antiepileptogenic treatments. In many acquired epilepsy cases, there is a latent period, i.e., epileptogenesis (EPG), between the brain injury and the onset of spontaneous recurring seizures. During this latent period, affected brain tissue is thought to transform such that it eventually can generate spontaneous seizures \cite{pitkanen2014past}. Over 30\% of the patients will be pharmacoresistant and continue to suffer from recurring seizures despite intake of medications \cite{kwan2000early}. The more seizure episodes have occurred before the first clinical visit, the less effective of the treatment will be \cite{kwan2000early}. Hence, identifying the presence of EPG before the epilepsy is fully established would be of great importance. However, the process of EPG is still not fully understood \cite{pitkanen2016advances}. The precise time of onset of the brain being epileptogenic is untraceable
	\cite{pitkanen2014past}. However, it is safe to say that any antiepileptogenic or disease-modifying therapies should be administered as early as possible \cite{loscher2019holy}.	
	Thus, discovering prominent features of EPG could facilitate early diagnosis and open the door for early interventions \cite{moshe2015epilepsy}.
	
	Electroencephalography (EEG) is a common tool in the clinic due to its non-invasive and easy-to-deploy properties. However, detecting EPG from EEG data is challenging. Two reasons are the complexity of the mechanisms of EPG and the immense cross-subject variability, which result in different phenotypes of EEG signals. This makes reliable interpretation of EEG signals from previously unseen individuals difficult.
	
	Some works have attempted to identify electrophysiological biomarkers of EPG based on various hand-selected features \cite{bentes2018early}, \cite{rizzi2019changes}, \cite{milikovsky2017electrocorticographic}, \cite{bragin2004high}, \cite{bragin2016pathologic}. However, a manual selection of features may be biased and overlook useful information.
	Recently, fueled by advances in ML, impressive results have been achieved in a variety of domains by training on raw data and letting the learning algorithm identify useful features automatically. Such approaches can even outperform human experts \cite{hannun2019cardiologist}, \cite{haenssle2018man}, \cite{sarker2018slsdeep}.
	
	Here, we recorded intracranial EEG signals from a rodent model of mesial temporal lobe epilepsy with hippocampal sclerosis (mTLE-HS) \cite{norwood2010classic}. In this model, epilepsy was induced by electrical perforant pathway stimulation (PPS) through depth electrodes. Continuous EEG recordings were obtained from the hilus of the dentate gyrus after the implantation of the electrode until the occurrence of the first spontaneous seizure. The EEG recordings were divided into two classes depending on the time of recording relative to the PPS stimulation. The samples recorded before the epilepsy-triggering PPS form the {\em baseline} (BL) class. The samples recorded after the PPS, but before occurrence of the first spontaneous seizure form the {\em epileptogenesis} (EPG) class. In the following, we propose a deep learning (DL) framework to classify EPG vs. BL by training on raw EEG data in an end-to-end fashion.
	
	To tackle the problem that a large portion of normal brain EEG patterns are also present in the EPG phase, we propose a prediction aggregation method where predictions from a longer time interval (e.g.\ one hour) are pooled together through a linear aggregation. We assume that the ``EPG-typical" signals should be more frequent during the EPG phase compared to the baseline phase. This difference becomes apparent through the aggregation method. Specifically, we make the following contributions:
	
	
	\begin{itemize}
		\item we present the first attempt to identify the process of EPG with a deep neural network (DNN) trained on EEG time series data. This is a radical departure from the conventional (and hitherto not very successful) approach of attempting to predict individual seizures when the disease has already established itself.
		\item  We propose a framework for EPG identification using massive amounts of EEG data from chronic recordings to maximally exploit the DNN's learning pability and minimize human effort in data labeling and feature engineering.
		\item  We use a prediction aggregation method and demonstrate that it achieves high fidelity EPG detection in a rodent model.
		\item We show that just two minutes of recordings can be sufficient to achieve reasonably good classification performance.
	\end{itemize}
	
	\subsection*{Generalizable Insights about ML in the Context of Healthcare}
	Massive expert annotations are expensive and therefore often scarce in medical contexts. This poses tremendous difficulties for the application of ML. When large amounts of data can be collected but labelling by experts is infeasible, turning to a form of ``cheap" labelling can be a way-out. In our case, detailed expert annotations are absent but the EEG signals are recorded continuously (24/7), which yields a large quantity of training data. We define the labels exclusively according to the relative time of the recording with respect to the PPS. This kind of label is cheap and easy to obtain but less informative, since in the EPG period large amounts of normal brain activity are still present, i.e., the data from the two classes are largely overlapping. To deal with this large overlap, we propose a prediction aggregation process to pool decisions over a long time window. We show here that even in the complete absence of expert annotations of specific events showing ``EPG-typical" brain activity, the large data set in combination with the ``cheap" labels allow us to build a powerful classification system. We suggest that many other medical problems where the application of ML is currently infeasible due to lack of detailed expert annotations could be tackled using similar methods. More generally, our approach of massive data collection to identify the earliest signatures of a developing disease may enable early diagnosis and intervention across a wide range of medical contexts.

	\section{Related Work}
	\paragraph{EEG Analysis with DL}
	Modern ML techniques allow an end-to-end learning approach to the analysis of EEG data rather than relying on specific handcrafted features. In particular, DNNs have been applied to either frequency representations \cite{lu2019deep}, \cite{thodoroff2016learning} or directly to raw EEG data in the time domain \cite{kiral2018epileptic}, \cite{biswal2019eegtotext}, \cite{avcu2019seizure}, \cite{farahat2019convolutional}, \cite{bi2019early}. They have achieved promising results in seizure detection, seizure prediction, or even other neurological disorders such as Alzheimer's disease and Autism classification. For example,
	Zhou \textit{et al.} compared the performance of a CNN on the EEG signal classification problem with time-domain and frequency-domain input and concluded that frequency-domain signals have greater potential for the task \cite{zhou2018epileptic}.
	Kiral-Kornek \textit{et al.} demonstrated an accurate, automated
	patient-specific seizure prediction approach with a DNN trained on intracranial EEG data \cite{kiral2018epileptic}.
	Biswal \textit{et al.} applied stacked CNNs and recurrent neural networks (RNNs) to extract temporal shift invariant features from EEG data \cite{biswal2019eegtotext}. These features are used to classify multiple key EEG phenotypes. 
	Avcu \textit{et al.} developed an end-to-end solution for seizure onset detection \cite{avcu2019seizure}.
	Bi and Wang applied a convolutional deep Boltzmann machine with EEG data in early diagnosis of Alzheimer's disease \cite{bi2019early}.
	Thodoroff \textit{et al.} applied a deep RNN with a CNN to perform automated patient specific seizure detection with scalp EEG \cite{thodoroff2016learning}. 
	A deep CNN is applied for EEG signal decoding during human decision making process and demonstrates promising results \cite{farahat2019convolutional}.
	
	\begin{figure}[tb]
		\centering
		\includegraphics[width=0.9\linewidth]{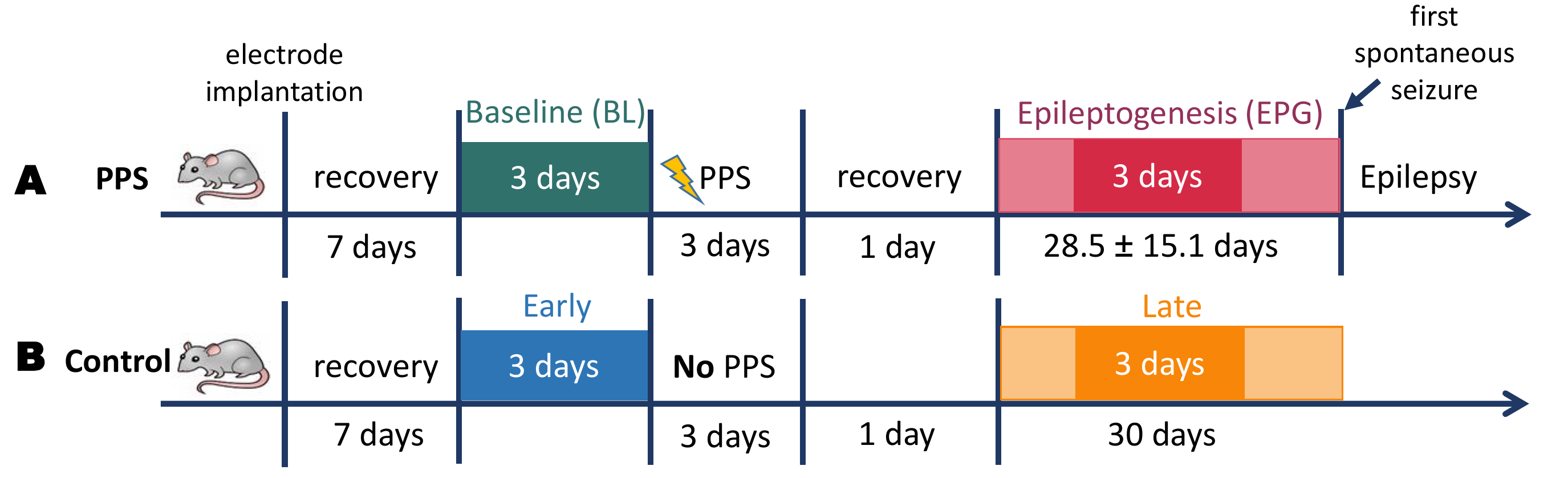}
		\caption{Timeline of the experiment. A. time line of PPS-stimulated rats. B: time line of control rats.
			PPS: perforant pathway stimulation.}
		\label{timespan}
	\end{figure}
	\paragraph{EPG Biomarker in EEG} There have been several previous studies on biomarker discovery for identifying EPG. Bragin \textit{et al.} found that the occurrence of high-frequency-oscillations (HFOs) is a strong indicator of future recurrent spontaneous seizures and the sooner HFOs occur, the shorter the EPG period will be \cite{bragin2004high}. Andrade \textit{et al.} found that a duration reduction of sleep spindles at the transition from stage III to rapid-eye-movement sleep indicates potential post-traumatic epilepsy in a lateral fluid-percussion rat model \cite{andrade2017generalized}.
	In humans, it was shown that over 90\% of the HFO area overlapped with the seizure onset zone for six patients \cite{burnos2014human}. 
	Milikovsky \textit{et al.} revealed that the dynamics of the theta band could predict future post-injury epilepsy and the seizure onset and thus could serve as a diagnostic biomarker for EPG \cite{milikovsky2017electrocorticographic}. 
	Lu \textit{et al.} demonstrated that an increased delta band power, a decreased of theta band power as well as an increase of high gamma band power were correlated with the presence of EPG in a rat mesial temporal lobe epilepsy model \cite{lu2019deep}.
	Rizzi \textit{et al.} recently showed using concepts from nonlinear dynamics, that a reduction of the dimensionality of EEG/ECoG signals indicates the presence and the severity of EPG in three different rodent epilepsy models \cite{rizzi2019changes}.
	Bentes \textit{et al.} found that an asymmetry in background EEG signals and interictal epileptiform discharges can independently predict the post-stroke epilepsy in a clinical study \cite{bentes2018early}. 
	However, so far a DL-based approach to EPG biomarker discovery has not yet been attempted.

	\section{Methods}
	
	\begin{figure}[tb]
		\centering
		\includegraphics[width=0.9\linewidth]{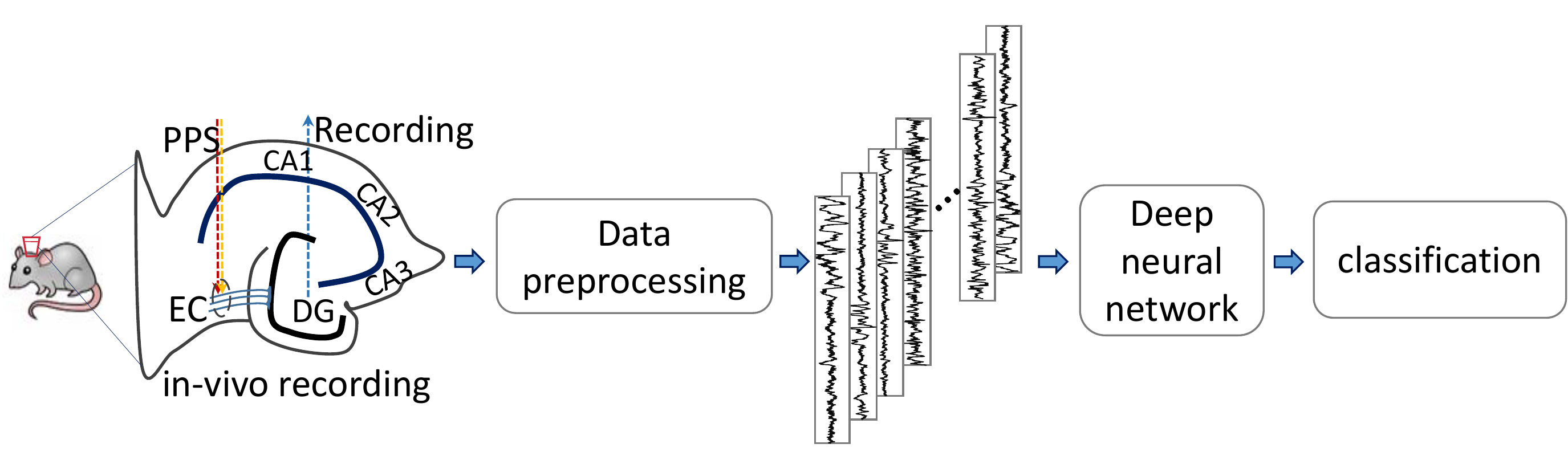}
		\caption{Workflow of our proposed framework. EC: entorhinal cortex, DG: dentate gyrus, CA: cornu ammonis, PPS: perforant pathway stimulation.}
		\label{workflow}
	\end{figure}
	
	\subsection{Dataset}
	
	\label{dataset}
	We used intracranial EEG data recorded continuously (24/7) by a depth-electrode from a rodent mTLE-HS model, where epilepsy is induced by electrical PPS, as described in detail in \cite{costard2019electrical}. The stimulated rats developed epilepsy after an average EPG phase of four weeks (range one to seven weeks). The EPG phase ended with the first spontaneous seizure. 
	
	The rat model provides an opportunity to study the progression of epilepsy and to discover potential biomarkers of EPG in the EEG. In this study, we included seven PPS-treated rats with continuous wireless EEG recordings. We also included three control rats which had electrodes implanted but did not undergo PPS and did not develop epilepsy by the end of the recording (the lifetime of battery of the wireless transmitter). The time-lines for the PPS group and the control group are shown in Fig.~\ref{timespan}. We denoted two phases of interest from the continuous recording, i.e., baseline (BL) and epileptogenesis (EPG). In our study, we selected the last three days from the BL phase and assign to them the label ``0". We selected the 7th, 8th and 9th day of the EPG phase for all rats in order to maintain the maximum time distance to acute symptomatic seizures which can occur within the first 1-3 days after PPS, while the first spontaneous seizure in our animal group appeared on day 10. The samples from this EPG period were assigned the label ``1". 
	
	\paragraph{Preprocessing}
	The sampling rate of the EEG recordings  was 512 Hz. A band-pass filter between 0.5 - 160 Hz and a notch filter at 50 Hz were applied to the raw data. In our experimental setting, the recorded EEG signals were susceptible to electric interference, which  results in high amplitude outliers. To fix this problem, we applied an outlier filtering method to get rid of unrealistic values in the recordings. We obtained non-overlapping five-second segments from the continuous recordings. To clean up the data for training, those segments with more than 20\% data loss due to weak wireless transmission were discarded. The workflow is shown in Fig~\ref{workflow}.

	\label{model}
	Our proposed method consists of two parts: (a) a deep residual neural network and (b) a prediction aggregation process during the testing.
	
	\paragraph{Residual convolutional neural network}
	
	Our model is a DNN with 33 convolutional layers with residual connections and it is inspired by the work of Hannun \textit{et al.} \cite{hannun2019cardiologist}. The network's structure is shown in Table~\ref{structure}. The concept of residual connections was first proposed by He \textit{et al.} \cite{he2016deep} for an image recognition task and has been widely used in a variety of tasks such as image segmentation \cite{huang2017densely}, \cite{lei2018temporal}, \cite{liu2019auto}, visual object detection \cite{mordan2018revisiting}, \cite{wang2019towards}, and healthcare-related applications \cite{hannun2019cardiologist}, \cite{sarker2018slsdeep}. The residual connection connects the pre-activation from one layer with the input of another previous layer in an additive fashion skipping several layers in between. Then, the non-linear activation is applied to the addition to compute the input for the next layer. The collection of the computations between one residual connection is termed a block. The output of the network is a softmax layer taking the flattened feature maps as input and outputs a probability distribution over two possible classes.
	
	In our work, the network structure is in the same spirit as in the work of Hannun \textit{et al.} \cite{hannun2019cardiologist} and we performed a hyper-parameter exploration for our specific task. A drop-out rate of 0.25 yields the best performance among the values 0.2, 0.25, 0.3, 0.5, and 0.65. The number of blocks that performs best is a value of 15 among 5, 7, 11, and 15. A filter size of 32 works the best among values of 3, 9, 11, 16, 32, and 64. We tried ReLU and leaky ReLU as the nonlinear activation function and no significant difference was observed, so we chose the ReLU activation for this work. A starting number of 16 filters yields better results than 8 and 32. 
	
	We adopted the pre-activation design from Hannun \textit{et al.} \cite{he2016identity}. The convolutional layer has a filter width of 32. The number of filters increases by a factor of 2 in every four blocks starting from 16. The feature maps were down-sampled in every other block with a stride of 2. To keep the dimensionality compatible, the max-pooling branch shares the same stride value as in the second convolutional layer in each block.

	\begin{table}[t]
		\caption{The network structure used in our work. The \textbf{Config} column show the filter size (always 32) and the number of filters we use in each convolutional layer. The number of filters is increased every four blocks by a factor of 2. Every other block subsamples its input by a factor of 2, indicated by the value of \textbf{stride}. Here, the batch size at the first dimension is omitted in the output shape column}
		\label{structure}
		\centering
		\scalebox{0.95}{
			\begin{tabular}{lcccc}
				\toprule
				\textbf{Name}     & \textbf{Config}      & \textbf{Stride} & \textbf{Factor i} & \textbf{Output shape}\\
				\midrule
				Conv layer 0 & $\left[ \begin{array}{cc} 32\times1, & 16\times 2^i \end{array}\right]$ & 1 &  0 & [2560, 1, 16] \\
				\midrule
				ResBlock 0 & $\left[ \begin{array}{cc} 32\times1, & 16\times 2^i \\ 32\times1, & 16\times 2^i \end{array}\right]$   & 1 &  0  & [2560, 1, 16]  \\
				\midrule
				ResBlock 1     & $\left[ \begin{array}{cc} 32\times1, & 16\times 2^i \\ 32\times1, & 16\times 2^i \end{array}\right]$ & 2 &  0   & [1280, 1, $16 \times 2^i$]     \\
				\midrule
				ResBlock 2     & $\left[ \begin{array}{cc} 32\times1, & 16\times 2^i \\ 32\times1, & 16\times 2^i \end{array}\right]$  & 1 &  0   & [1280, 1, $16 \times 2^i$]    \\
				\midrule
				ResBlock 3     & $\left[ \begin{array}{cc} 32\times1, & 16\times 2^i \\ 32\times1, & 16\times 2^i \end{array}\right]$  & 2 &  0   & [640, 1, $16 \times 2^i$]    \\
				\midrule
				ResBlock 4     & $\left[ \begin{array}{cc} 32\times1, & 16\times 2^i \\ 32\times1, & 16\times 2^i \end{array}\right]$  & 1 &  1 & [640, 1, $16 \times 2^i$]      \\
				\midrule
				ResBlock (5,\dots, 8)     & $\left[ \begin{array}{cc} 32\times1, & 16\times 2^i \\ 32\times1, & 16\times 2^i \end{array}\right]$  & (2, 1, 2, 1) &  (1, 1, 1, 2)  & [320, 1, $16\times 2^i$]    \\
				\midrule
				ResBlock (9,\dots, 12)     & $\left[ \begin{array}{cc} 32\times1, & 16\times 2^i \\ 32\times1, & 16\times 2^i \end{array}\right]$    & (2, 1, 2, 1) &  (2, 2, 2, 3) & [80, 1, $16\times 2^i$]   \\
				\midrule
				ResBlock (13, 14)    & $\left[ \begin{array}{cc} 32\times1, & 16\times 2^i \\ 32\times1, & 16\times 2^i \end{array}\right]$   & (2, 1) &  (3, 3)  & [20, 1, $16\times 2^i$]   \\
				\midrule
				Dense     &  2  &  &   & [2]  \\
				\bottomrule
			\end{tabular}
		}
	\end{table}

	\paragraph{Prediction aggregation}
	We hypothesize that the EPG phase may be better characterized by a change of distribution of different waveforms rather than a specific waveform that can be identified in every individual segment. Therefore a reliable classification can only be achieved by pooling information from many data segments. Our method is inspired by Smyth and Wolpert \cite{smyth1999linearly}. For each segment, the network outputs how likely this segment is taken from each class. Then, we linearly aggregate the predictions for multiple consecutive segments to obtain the final classification result. 
	
	Considering the data pairs, the EEG segments are $x_{(h, i)}$ and the associated labels are $y_{(h, i)}$ in one continuous hour $h$, where $i=1,\dots, N$ and $N$ is the total number of the samples in this hour. The softmax output of these samples is given by $\hat{y}_{(h, i)}=f(x_{(h, i)}, \text{model})$ and it is in shape $[N, 2]$ where 2 is the number of classes in our supervised scheme. The aggregated prediction for hour $h$ is given by $\hat{y}_h = \sum_{i=1}^{N}\hat{y}_{(h, i)}= \sum_{i=1}^{N}f(x_{(h, i)}, \text{model})$, 
	and in shape of $[1, 2]$. In a final step, we normalize $\hat{y}_h$ along the column axis. The resulting number is interpreted as a class probability and used to compute corresponding performance metrics.
	
	\paragraph{Training procedure}
	We apply leave-one-out (LOO) cross validation to test the generalization ability of our approach for both the PPS group and the control group. To be specific, in each group we iterate down the list of rat IDs holding out the data from the current rat for future test and only train and validate on the data from other rats. During training and validation, we randomly select 25 hours from each phase and from each rat, which yields in total around 213,000 training samples. We adopted a train-validation-split of 9:1. The choice of 25 hours is a good trade-off between computation cost and performance chosen empirically. After the network is trained, we test it with the previously withheld data, which contains over 100,000 test samples.

	\section{Experiments and Results}
	\label{experiment}
	\subsection{Experiments Design}
	To evaluate our method’s ability to identify EPG, we designed two tasks: task A is designed to classify BL vs.\ EPG signals in PPS rats as shown in Fig~\ref{timespan}A. Task B is a control designed to classify signals recorded in the early and late implantation phases in the set of control rats as shown in Fig~\ref{timespan}B. 
	
	\paragraph{Task A: BL vs.\ EPG classification in PPS rats}
	This is our main task in which we want to distinguish EEG signals from BL and EPG phases. In this task, we applied seven-fold LOO cross validation with the data from the seven PPS-stimulated rats. 	
	
	\paragraph{Task B: \textit{early} vs.\ \textit{late} classification in control rats}
	In this control task we want to rule out the possibility that differences between BL and EPG in Task A could be simply due to systematic changes in the tissue after electrode implantation that have nothing to do with the EPG triggered by PPS. Therefore, we study control rats that do not undergo PPS (see Fig~\ref{timespan}B) and analyze if there are systematic differences between the EEG signals recorded from the early and late implantation phases. We applied a three-fold LOO cross validation scheme with the same network configuration as in task A.
	
	\begin{table}
		\caption{Performance matrices without (5 second) and with one hour of aggregation. Data are presented
			as mean $\pm$ standard deviation. SEN: sensitivity, SPE: specificity, AUC: area under the curve}
		\label{1-hour}
		\centering
		\begin{tabular}{cccccc}
			\toprule
			\textbf{Aggregation length} & \textbf{Task}     & \textbf{SEN} & \textbf{SPE} & \textbf{AUC} \\
			\midrule
			& Task A & $0.73 \pm 0.25$ & $0.77\pm0.17 $&  $0.86\pm0.07$ \\
			5 second &Task B  &$ 0.57\pm0.42$ &   $0.43\pm0.42$ & $0.50\pm0.08$  \\
			\midrule
			& Task A  & $\textbf{0.94} \pm 0.05$ & $\textbf{0.96}\pm0.04 $& $\textbf{0.99}\pm0.01$\\
			1 hour & Task B  &$ 0.63\pm0.45$ &   $0.37\pm0.45$ & $0.45\pm0.06$  \\
			\bottomrule
		\end{tabular}
	\end{table}
	
	\begin{figure}[th]
		\centering
		\includegraphics[width=0.99\linewidth]{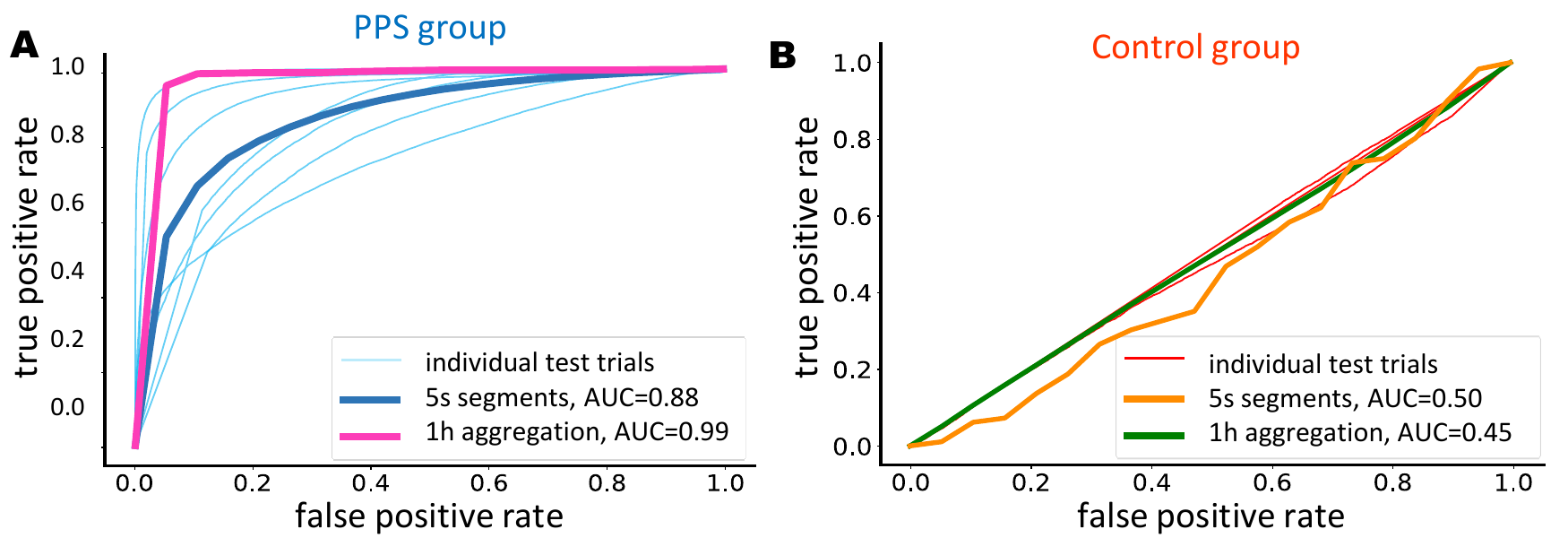}
		\caption{Receiver operating characteristic (ROC) curves. \textbf{A}: \textbf{The PPS group} (seven rats). Individual ROC curves from all LOO test trials (thin light blue), the average ROC curve without prediction aggregation (thick blue) and the average ROC curve with aggregation in a continuous stretch of one hour (thick pink). \textbf{B}: \textbf{The control group}. Individual ROC curves (thin light orange), the average ROC curve without aggregation (thick orange) and the average ROC curve with aggregation over one hour (thick green) from all LOO test trials in the control group. AUC: area under the curve. PPS: perforant pathway stimulation. LOO: leave-one-out}
		\label{ROCs}
	\end{figure}

	\subsection{Results} 
	\subsubsection{ROC Analysis}
	The average ROC curves of all the leave-one-out test trials in each task are shown in Fig~\ref{ROCs}. The AUC values are computed in two scenarios: a) each five second segment is viewed independently and the AUC is calculated based on the prediction of all the five second segments, b) the predictions of multiple consecutive five second segments are aggregated together through a linear stacking. In Fig~\ref{ROCs}A, we show the ROC curves in individual LOO test trials, and the averaged ROC curves with and without prediction aggregation. Our method could discern signals from both phases with an average AUC under the ROC curve of 0.88. It suggests that the neural network has learned features that are informative for the correct classification. With the proposed prediction aggregation over one hour, the average AUC achieves 0.99, which shows that the proposed approach can reliably discern EEG signals from the BL and the EPG phase. In contrast, for the control group, the \textit{early} vs.\  \textit{late} phase classification, the network does not show clear discriminative ability. The average AUCs from all the test trials with and without the prediction aggregation are 0.50 and 0.45, respectively. The detailed performance measurements such as sensitivity (SEN)~=~$\frac{TP}{TP + FN}$, specificity (SPE)~=~$\frac{TN}{TN + FP}$ and the AUC are shown in Table.~\ref{1-hour}, where \textit{TP, TN, FP, FN} denote true positive, true negative, false positive and false negative, respectively.
	
	\begin{figure}[tb]
		\centering
		\includegraphics[width=0.99\linewidth]{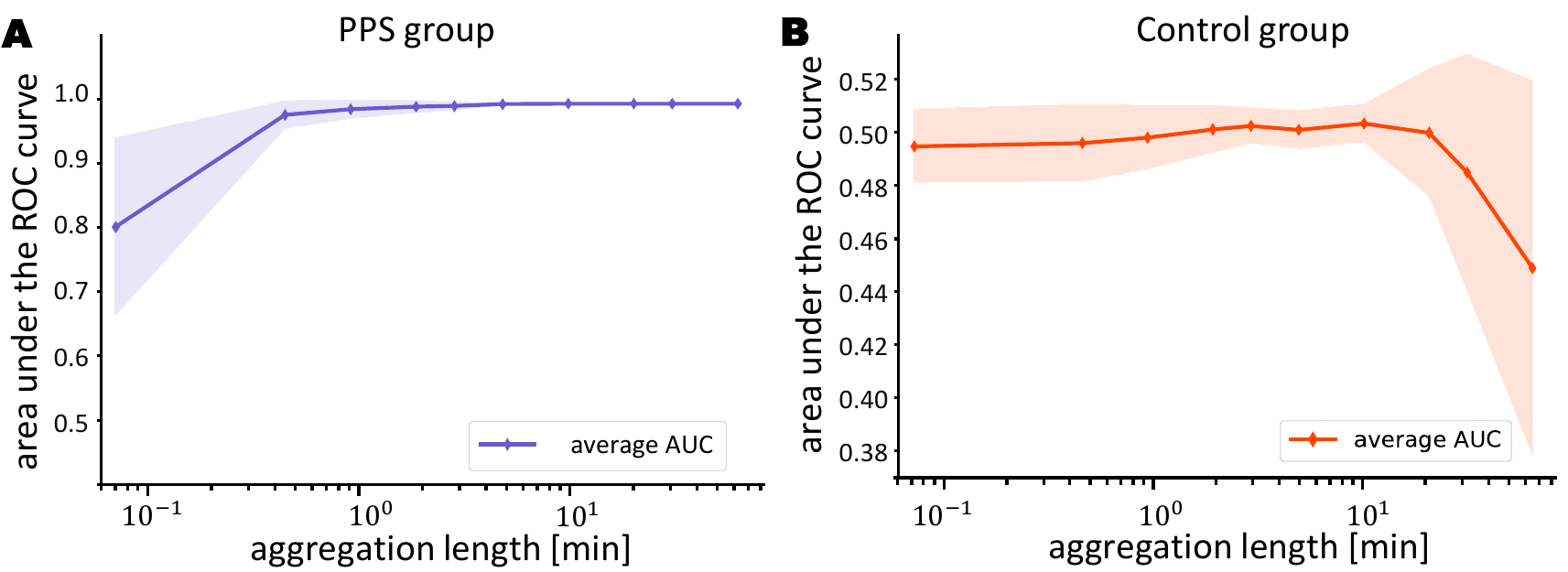}
		\caption{The average AUC over all leave-one-out test trials as a function of the aggregation length for the two groups. The shaded area represents one standard deviation.}
		\label{average-AUC}
	\end{figure}

	\subsubsection{Aggregation Effect}
	\begin{figure}[bt]
		\centering
		\includegraphics[width=0.99\linewidth]{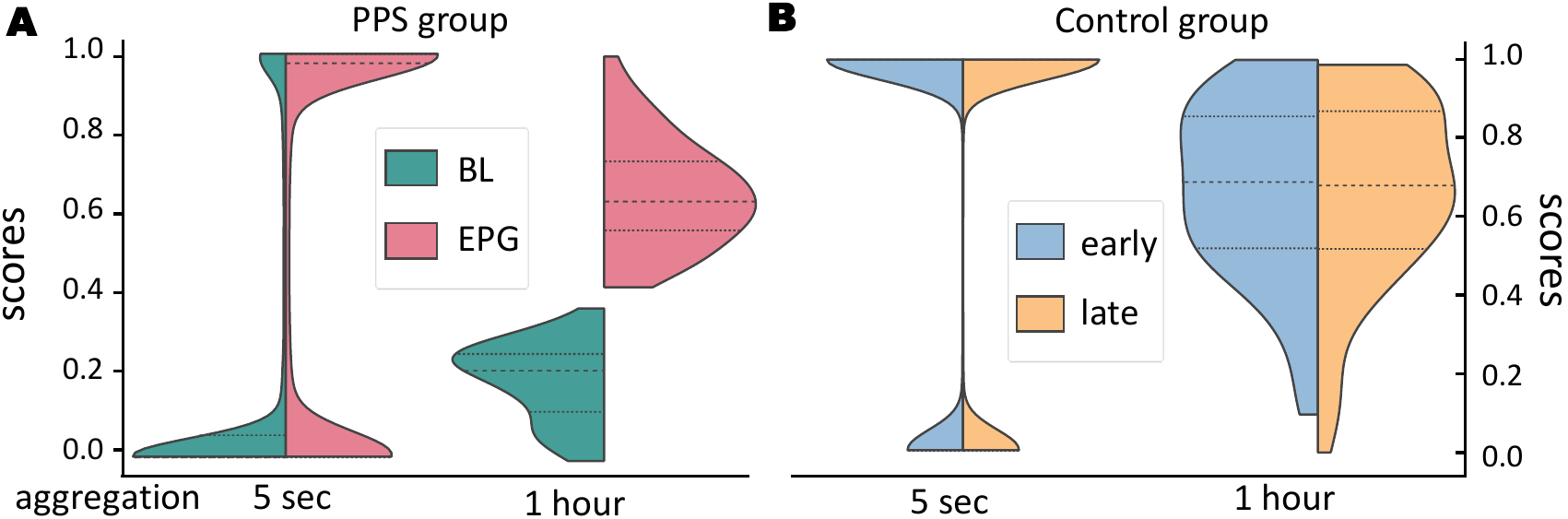}
		\caption{Example distributions of scores from both classes.  \textbf{A}: \textbf{PPS group}. \textbf{(left)} without aggregation. The mean and variance of the two distributions, i.e., from all BL segments and all EPG segments, are different but overlapping. \textbf{(right)} with one hour of aggregation.  \textbf{B}: \textbf{Control group}. \textbf{(left)} without aggregation. \textbf{(right)} with one hour of aggregation. BL: baseline. EPG: epileptogenesis}
		\label{violin}
	\end{figure}
	To further investigate the effect of aggregation, we computed the AUC value in each test trial with various intervals, i.e., five seconds, 30 seconds, one, two, five, ten, 20, 30, 60 minutes. The average AUC across all the test trials in the PPS group as a function of the aggregation lengths is shown in Fig~\ref{average-AUC}A. It shows a clear trend of an increasing AUC and a decrease of standard deviation with a longer aggregation length. Thus, the prediction aggregation from multiple consecutive segments is essential for a strong performance in the PPS group. In contrast, in the control group, the aggregation not only did not help increase but reduced the average AUC, as depicted in Fig~\ref{average-AUC}B. The result also suggests that two minutes of continuous EEG recordings are sufficient to achieve an AUC value above 0.95.
	
	We also tested if the seven neural networks trained on the PPS group would discriminate the \textit{early} and \textit{late} phase EEG patterns from the control animals. If so, this would suggest that these networks learn to discover changes in the EEG patterns across time that are triggered by the surgical procedure but are independent of the PPS and the ensuing EPG. However, we found that these networks could not discriminate \textit{early} and \textit{late} EEG patterns from the control group (mean AUC = 0.53, std. dev. = 0.12) and over 82\% of all test samples from both \textit{early} and \textit{late} phases are classified as BL. This is additional evidence that the networks have learned to detect changes in EEG patterns that are induced by the PPS.
	

	To visualize how exactly the prediction aggregation improves the discriminative ability of the model, we compute the distribution of scores assigned by the network to all test segments. Notably, the \textbf{score} is defined as the softmax output of the segment being EPG. Ideally, scores for BL segments should be close to zero, and EPG segments should have close-to-one scores. For simplicity, we only show the distributions of one LOO test trial from each group, as presented in Fig~\ref{violin}. The difference of the distributions within the same aggregation length is evaluated with the ANOVA test and the Wilcoxon rank sum test. In Fig.~\ref{violin}A, the distributions are significantly different in both cases for this rat (the ANOVA test, p-value $\le 10^{-25}$, the Wilcoxon rank sum test, p-value $\le 10^{-17}$). It shows the same trend and the statistical values are at the same magnitude for all the other PPS-stimulated rats. Since the number of samples is sufficiently large, the p-values are very small. To measure the sizes of differences between two distributions within the same aggregation length, we also computed Cohen's \textit{d} effect size \cite{rice2005comparing}. In two aggregation cases in the example test trial, the \textit{d} = 0.94 and 2.91, respectively. Average \textit{d} values for the whole PPS-stimulated group with and without aggregation are 0.85 and 1.24, respectively. Cohen suggested that an effect size absolute value over 0.8 is considered large. Notably, there is still a considerable overlap between BL and EPG segments, i.e., in BL period there are a certain number of segments classified as EPG and vice versa. When we aggregate over one hour, the effect of the distribution shift is magnified. 
	In contrast, in the control group, the distributions of scores from one test trial with and without aggregation, are shown in Fig~\ref{violin}B, are not significantly different (the ANOVA test, p-values $\ge 0.5$, the Wilcoxon rank sum test, p-values $\ge 0.4$) with an effect size \textit{d} = 0.004 and 0.012, respectively. The other two LOO test trials in the control group exhibit the same pattern.
	
	\begin{figure}[t]
		\centering
		\includegraphics[width=0.99\linewidth]{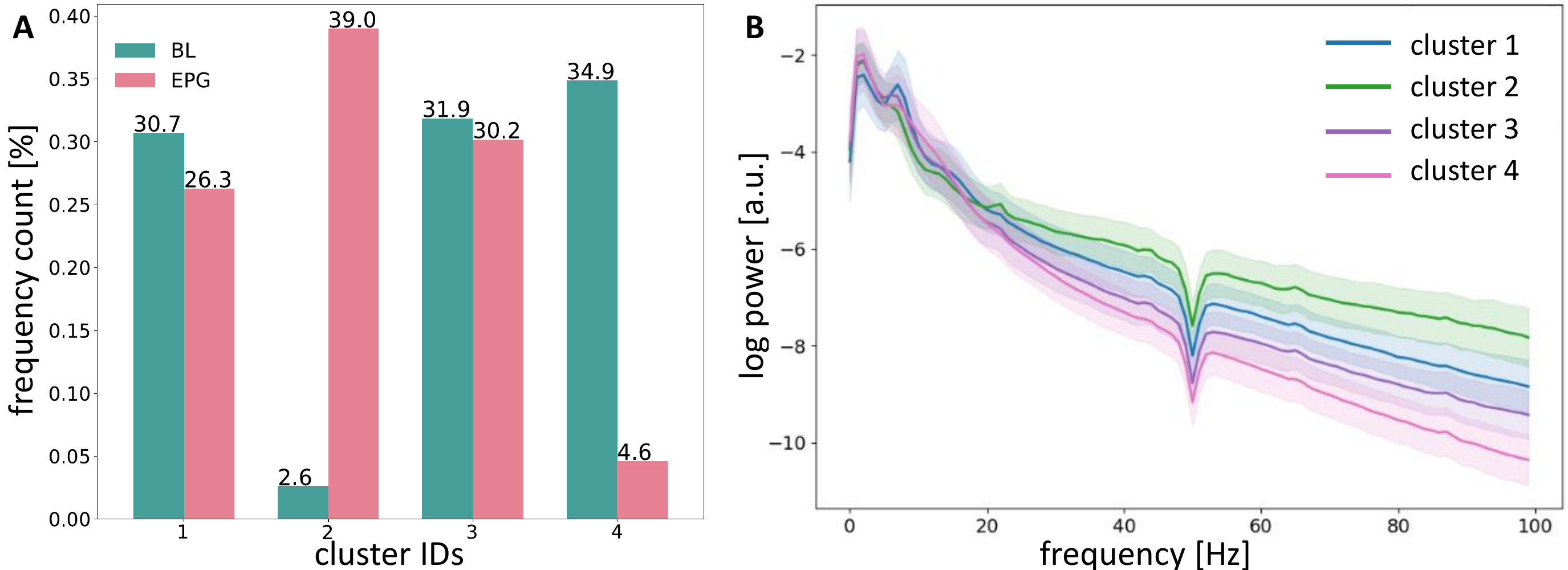}
		\caption{Clustering of high EPG score examples. \textbf{A}: percentage count of each class in each cluster. \textbf{B}: mean spectra of each cluster. The shaded area represents the standard deviation. }
		\label{cluster}
	\end{figure}

	\subsubsection{K-Means Clustering Analysis}
	In order to obtain a better understanding of the characteristics of the learned features, we conducted k-means clustering analysis on very certain samples collected from LOO test-trials. Here, a certain sample is defined as one whose softmax probability is larger than a threshold (set to 0.999). The k-means clustering analysis is based on the Euclidean distance. We cluster the log-power spectrum of examples into four clusters, where a number of four is determined by the elbow-theory \cite{kodinariya2013review}. From the frequency count plot, see Fig.~\ref{cluster}A, we can see that the majority of the cluster No.~2 is from the EPG class and that of the cluster No.~4 is from BL class. From the mean spectra of each cluster, we can see that the EPG-dominant cluster has higher power in the frequency range over 20 Hz to 100 Hz. Specially, in this cluster, there is strong power around 22 Hz and its harmonics. On the other hand, the mean power spectrum of the BL-dominant cluster, cluster No.~4, has a faster decay towards higher frequencies.
	
	\section{Discussion}\label{discussion} 
	
	In recent years, ML could capitalize on the availability of big medical data sets. However, acquiring expert annotations for such data is impractical in many applications, representing a challenge for ML approaches. Here, we have tried to answer the question if an emerging epilepsy might be detectable from EEG signals even before the first seizure occurs. For this, we have used a rodent model of epilepsy \cite{costard2019electrical}, where epileptogenesis (EPG) is triggered through PPS. While massive amounts of training data are available from the BL (pre-stimulation) and the EPG (post-stimulation) periods, these data are only labeled by their time of recording. On one hand, there might be large amounts of EPG-like signals present in the BL phase because there is brain injury involved in implanting the electrode. On the other hand, normal brain activities are still present in the EPG phase. Thus, we can expect short segments of EEG recordings to be often indistinguishable. A reliable classification requires pooling data over longer time windows. To achieve this, we have proposed a DNN approach with a prediction aggregation method. Our method is trained in an end-to-end fashion on five second segments and we have observed massive performance gains when aggregating predictions over one hour (improvements of 21\%, 19\%, and 13\% in SEN, SPE, and AUC, respectively). Therefore, we  have demonstrated a very viable method for automatically predicting epilepsy from EEG recordings prior to the first epileptic seizure. This opens the door for early interventions to modify or even arrest the progression of the disease \cite{loscher2019holy}. Furthermore, EEG patterns that the network has identified as being predictive of EPG may point towards new biomarkers of the disease. As a plausible alternative approach to our network architecture, a recurrent neural network (RNN) could be considered. However, our preliminary investigations have shown that RNN training requires more structure exploration and hyper-parameter search and our results leave little room for improvement on the data set presented here. 
	
	\paragraph{Limitations}
	From the perspective of practical utility, a good biomarker for identifying EPG in a clinical setting should be noninvasive. In contrast, the data in our study were recorded with a depth electrode, which has a much higher signal-to-noise-ratio compared to surface EEG recordings. For training a similar model to predict EPG in humans, the collection of surface EEG data from human patients would be necessary. As an immediate next step, we plan to extend our results to a group of human patients, who will undergo EEG (surface or intracranial) recording in the hospital after suffering a brain injury but before epilepsy is manifest. With sufficient training data from these and non-epileptic patients, we could envision a machine-learning-assisted diagnostic tool for the early detection of a developing epilepsy in human patients.
	
	%
	%
	%
	\bibliographystyle{splncs04}
	\bibliography{biblio}
\end{document}